\documentclass{ifacconf}

\usepackage{graphicx}
\usepackage{amsmath}
\usepackage{natbib}
\usepackage{geometry}
\usepackage{url}
\begin{document}

\begin{frontmatter}

\title{Optimal Multi-Debris Mission Planning in LEO: A Deep Reinforcement Learning Approach with Co-Elliptic Transfers and Refueling}

\author[First]{Agni Bandyopadhyay}
\author[Second]{Günther Waxenegger-Wilfing}

\address[First]{Department of Computer Science, Julius-Maximilians-Universität Würzburg, Germany (e-mail: agni.bandyopadhyay@uni-wuerzburg.de)}
\address[Second]{Department of Computer Science, Julius-Maximilians-Universität Würzburg, Germany (e-mail: guenther.waxenegger@uni-wuerzburg.de)}

\begin{abstract}
This paper addresses the challenge of multi-target active debris removal (ADR) in Low Earth Orbit (LEO) by introducing a unified co-elliptic maneuver framework that combines Hohmann transfers, safety ellipse proximity operations, and explicit refueling logic. We benchmark three distinct planning algorithms—Greedy heuristic, Monte Carlo Tree Search (MCTS), and deep reinforcement learning (RL) using Masked Proximal Policy Optimization (PPO)—within a realistic orbital simulation environment featuring randomized debris fields, keep-out zones, and $\Delta V$ constraints. Experimental results over 100 test scenarios demonstrate that Masked PPO achieves superior mission efficiency and computational performance, visiting up to twice as many debris as Greedy and significantly outperforming MCTS in runtime. These findings underscore the promise of modern RL methods for scalable, safe, and resource-efficient space mission planning, paving the way for future advancements in ADR autonomy.
\end{abstract}

\end{frontmatter}

\section{Introduction}

The proliferation of space debris in Low Earth Orbit (LEO) has become a growing concern for the global space community, threatening both operational spacecraft and the long-term sustainability of space activities~\cite{ESA2023, NASA2008}. Over 36,000 trackable debris objects larger than 10 cm and millions of smaller fragments are now in orbit, resulting from collisions, break-ups, and routine operations~\cite{ESA2023, Kessler1978}. This increasing population, if left unmanaged, poses the risk of the so-called “Kessler Syndrome”---a cascading effect of collisions generating ever more debris~\cite{Kessler1978, Stenger2002}.

Active Debris Removal (ADR) is now widely recognized as an essential measure to ensure the safety and sustainability of space operations~\cite{Bonnal2013}. ADR missions typically involve a chaser spacecraft executing a sequence of autonomous rendezvous maneuvers to capture and de-orbit multiple debris targets within stringent fuel and time constraints~\cite{Federici2019}. Efficient planning of the visitation sequence is crucial to maximize mission effectiveness while minimizing resources.

Traditionally, the multi-debris rendezvous problem has been approached as a variant of the Traveling Salesman Problem (TSP)~\cite{Gutin2008, Federici2019}, using classical heuristics such as greedy search, genetic algorithms, or simulated annealing~\cite{Cormen2009, Halim2019}. While these techniques are computationally efficient and well-studied, they often fail to capture the full complexity of space mission constraints, such as refueling, keep-out zones, and dynamic collision avoidance.

Recent years have witnessed a surge in learning-based approaches for trajectory optimization and space mission planning. Deep reinforcement learning (RL), in particular, offers the ability to synthesize adaptive, scalable policies for high-dimensional, uncertain environments~\cite{Mazyavkina2021,Sutton2018,stephenson2025inspection,dunlap2024multiagent}. Notably, RL agents have demonstrated strong performance in sequencing debris rendezvous operations~\cite{Bandyopadhyay2024revisiting, Bandyopadhyay2025optimizing}, often outperforming popular traditional heuristics in both solution quality and computational speed.

In our previous work, \cite{Bandyopadhyay2024revisiting}, we introduced a policy gradient Reinforcement Learning (RL) agent for sequencing multi-debris visits, achieving a significant reduction in total rendezvous time compared to standard heuristics. We subsequently extended this framework to incorporate refueling and dynamic collision avoidance zones~\cite{Bandyopadhyay2025optimizing}, moving closer to the requirements of real-world ADR missions.

This paper advances the state of the art by introducing a unified co-elliptic maneuver framework that seamlessly integrates Hohmann transfers, safety ellipse detours~\cite{nasaatm2010}, and explicit refueling logic. To rigorously evaluate our approach, we benchmark our trained reinforcement learning agent against established heuristic baselines, namely Monte Carlo Tree Search (MCTS)~\cite{browne2012survey} and a classical Greedy algorithm~\cite{cormen2009introduction}. The mission planning challenge addressed in this work is inherently twofold: efficiency can be improved either through optimizing the underlying trajectory planning or by enhancing the decision-making strategy. In this study, we seek to deepen our understanding of the problem by making targeted advancements in both aspects.

\section{Related Work}

Multi-target trajectory optimization for active debris removal (ADR) has evolved from early combinatorial heuristics---such as the beam search, and greedy algorithms---to data-driven and reinforcement learning (RL) frameworks. Classical methods focused on minimizing cumulative $\Delta V$ or total transfer time across multiple targets~\cite{Letizia2017optim,Petit2012multiple,schaub}. Additionally, we incorporate real‐world constraints—namely fuel availability and overall mission duration—into the framework.

\begin{figure}[!t]
    \centering
    \includegraphics[width=\linewidth]{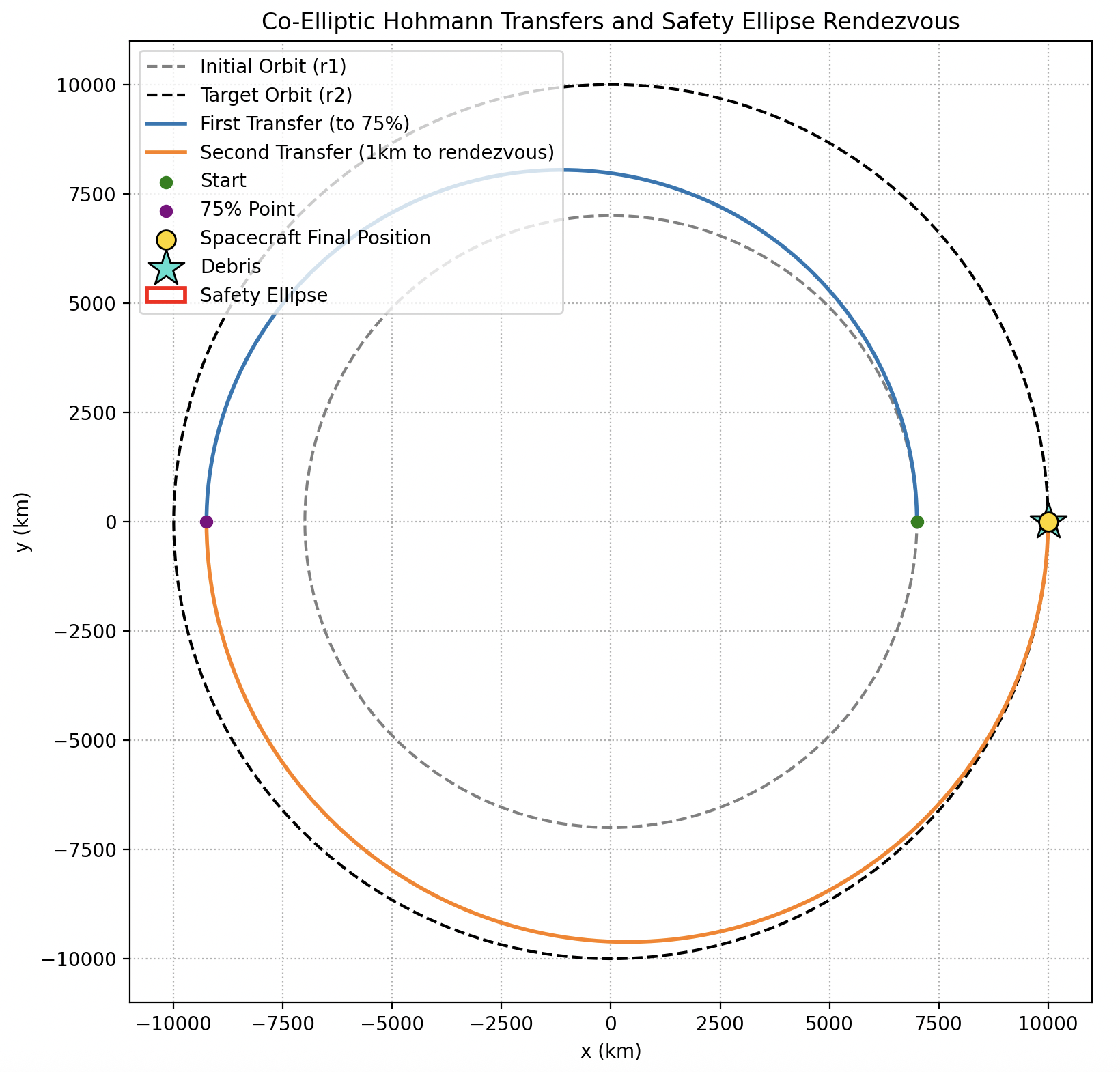}
    \caption{Overview of the co-elliptic Hohmann transfer sequence, showing the initial and target orbits, transfer maneuvers, and the terminal safety ellipse maneuver at rendezvous.}
    \label{fig:coelliptic}
\end{figure}

A fundamental maneuver for orbital rendezvous is the Hohmann transfer, which computes the minimum-energy transfer between two coplanar circular orbits~\cite{hohmann1925bestimmung,curtis2013orbital}. However, when visiting a sequence of debris objects distributed along similar orbits, performing a standard Hohmann transfer between each pair can lead to excessive and inefficient trajectory changes. To address this, \emph{co-elliptic} Hohmann transfers are used: the chaser spacecraft is placed into an intermediate co-elliptic orbit that shares the target's apogee or perigee and thus gradually phases around the Earth to rendezvous with the debris~\cite{vallado2001fundamentals}. This strategy allows for more efficient multi-target rendezvous when the debris are clustered in similar orbital bands, reducing both $\Delta V$ and mission time compared to a sequence of isolated transfers.It also significantly improves and quickens the phasing maneuvers and also leaves room for quick changes if necessary. Figure~\ref{fig:coelliptic} provides an overview of this co-elliptic maneuver sequence, highlighting the initial and target orbits, transfer maneuvers, and the terminal safety ellipse approach.

\begin{figure}[!t]
    \centering
    \includegraphics[width= 0.9\linewidth]{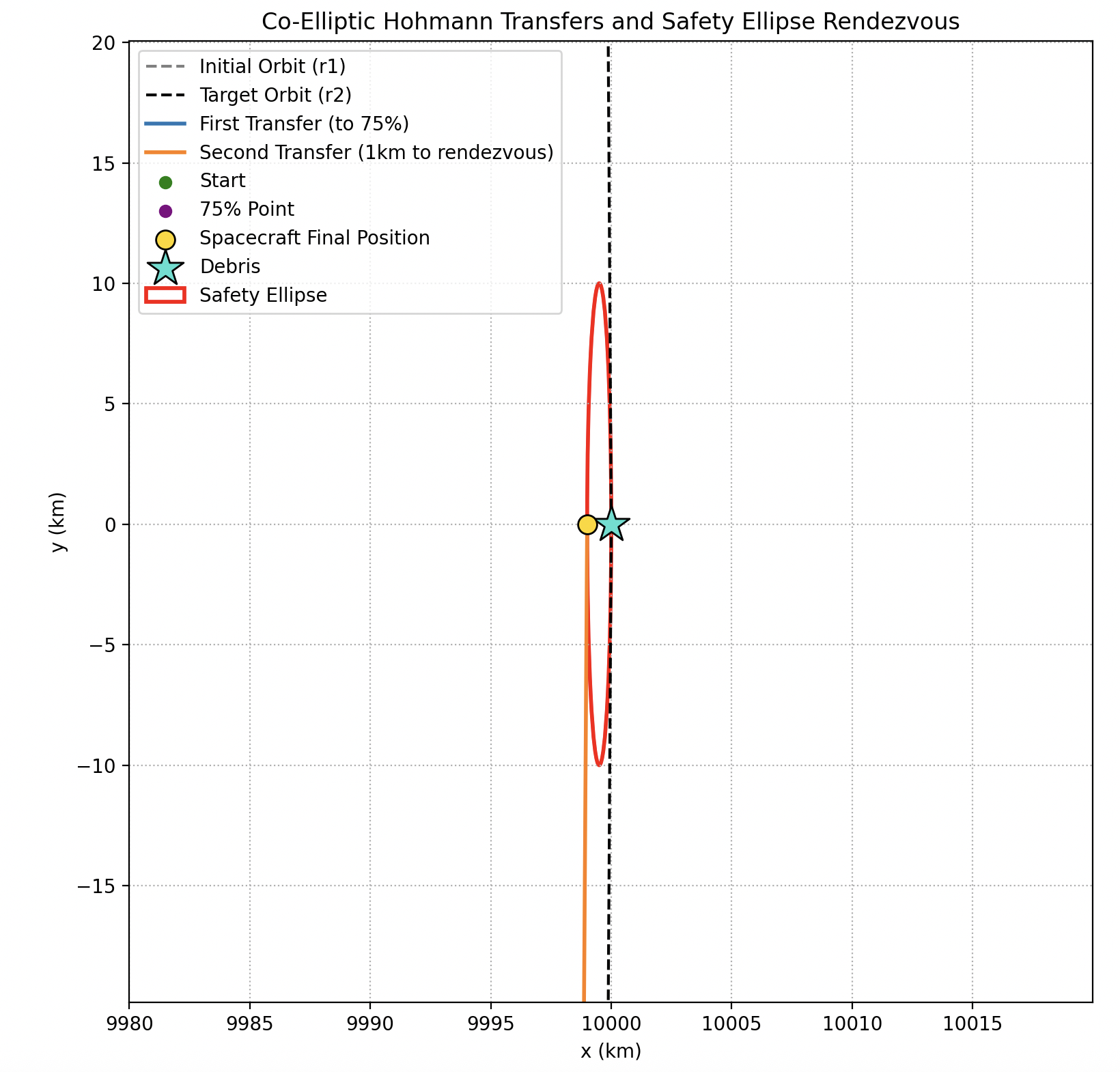}
    \caption{Zoomed-in view of the terminal approach phase, highlighting the safety ellipse maneuver for controlled and safe rendezvous with the target debris.}
    \label{fig:safetyellipse}
\end{figure}

Another critical aspect of proximity operations is the terminal approach and safety management. Standard Hohmann transfers do not address the need for controlled, precise, and safe approaches to non-cooperative targets like space debris. To mitigate the risk of collision during close-approach, recent frameworks integrate a \emph{safety ellipse maneuver} as proposed in~\cite{nasaatm2010}, which prescribes an elliptical slow approach to the target. This maneuver enables the chaser to approach within a predefined miss distance, maintaining both safety and controllability in the final rendezvous phase. Such manevers are widely adopted in operational spacecraft proximity operations to enforce strict safety requirements and provide margins against navigation errors and unmodeled dynamics~\cite{alonso2009collision}. Figure~\ref{fig:safetyellipse} illustrates a zoomed-in view of the terminal approach phase, showing the chaser spacecraft and debris within the safety ellipse during final approach.

Learning-based methods provide additional flexibility in this setting. Nevertheless, integrating co-elliptic transfers and safety ellipse maneuvers into sequential multi-target planning for ADR remains a challenging problem for both traditional and RL-based systems.

\section{Mission Model}

\subsection{Operational Scenario}

We consider an autonomous service spacecraft tasked with maximizing the number of debris objects visited and actively serviced within a fixed low Earth orbit (LEO) mission window. At the start of each episode, the chaser is initialized docked at a refueling station in a circular orbit of 700~km altitude, representing a typical LEO servicing platform. The environment generates a population of $N = 50$ debris targets per episode, each with randomly sampled altitudes, inclinations, and other orbital elements, reflecting the diversity and spatial spread of real-world LEO debris fields.

The agent operates in discrete decision steps. At each step, it may select a debris object for rendezvous or opt to return to the refueling station. Each debris visitation comprises a transfer maneuver, a brief period of station-keeping (e.g., one orbital period) to simulate servicing or data collection, and a subsequent decision to continue the mission or to refuel. Debris objects already visited are flagged and cannot be re-selected within the same episode.

\subsection{Transfer and Maneuver Model}

Inter-debris transfers are executed using a co-elliptic Hohmann transfer sequence to ensure realistic and efficient phasing and rendezvous operations. Specifically, for each target selection:
\begin{enumerate}
    \item The chaser initiates a maneuver to raise or lower its orbit to a co-elliptic trajectory that matches the altitude of the target debris, optimizing both transfer time and $\Delta V$.
    \item After appropriate phasing, a final burn circularizes the chaser's orbit to precisely match the debris' position, enabling rendezvous.
\end{enumerate}
The transfer sequence accounts for orbital inclination changes as needed, and ensures that all maneuvers adhere to the constraints of orbital mechanics.

The first Hohmann trasnfer is conducted to cover 75 percent of the distance between the chaser and the debris. Then the next Hohmann transfer brings the chaser within 1 km of the target after which it embarks on the safety ellipse maneuver to approach the target. The safety ellipse provides both control and accuracy in proximity operations.

\subsection{Refueling and $\Delta V$ Budgeting}

Propellant consumption is modeled directly via cumulative $\Delta V$ expended by the chaser, treating $\Delta V$ as a proxy for fuel usage. The agent may return to the refueling station at any time after successfully visiting at least one debris object. Upon docking, the agent's available $\Delta V$ budget is fully replenished; however, this incurs additional time and maneuvering penalties associated with the round-trip. This refueling logic encourages efficient path planning and penalizes excessive refueling cycles.

\subsection{Mission Constraints}

The following operational constraints are strictly enforced throughout each episode:
\begin{itemize}
    \item \textbf{$\Delta V$ Limit:} The chaser is initialized with a fixed $\Delta V$ budget (e.g., 3~km/s), which governs the maximum total velocity change available for all maneuvers before requiring refueling.
    \item \textbf{Mission Duration:} Each mission is bounded by a maximum allowed duration, such as 7 days, after which the episode terminates regardless of the number of debris visited.
    \item \textbf{Population Size:} Each episode contains a fixed set of $N = 50$ debris objects, with random orbital parameters to ensure diversity and robustness in planning strategies.
\end{itemize}

This operational scenario and maneuver model capture the critical challenges of multi-target rendezvous in cluttered LEO environments, while providing a flexible framework for benchmarking planning algorithms under realistic resource and safety constraints.

\section{Planning Algorithms}

This work benchmarks three distinct planning approaches for multi-target debris removal: a classical greedy heuristic, a Monte Carlo Tree Search (MCTS) planner, and a reinforcement learning (RL) agent. All algorithms operate under the same mission constraints and orbital mechanics, as described in Section~2.

\subsection{Greedy Heuristic}

The greedy algorithm selects, at each decision step, the next unvisited debris object $d^*$ that minimizes the incremental cost, defined as a weighted sum of $\Delta V$ and transfer time:
\begin{equation}
d^* = \arg\min_{d \in \mathcal{D}_{\text{unvisited}}} \left[ \alpha \Delta V_{\text{curr} \rightarrow d} + \beta T_{\text{curr} \rightarrow d} \right]
\end{equation}
where $\mathcal{D}_{\text{unvisited}}$ is the set of all remaining debris, $\Delta V_{\text{curr} \rightarrow d}$ is the velocity cost, $T_{\text{curr} \rightarrow d}$ is transfer time, and $\alpha, \beta$ are normalization or weighting coefficients (typically $\alpha=1$, $\beta=0$ for pure $\Delta V$ minimization). The greedy agent triggers a refueling maneuver if no reachable debris remains within the current $\Delta V$ and time budgets. This approach is inherently myopic, optimizing for immediate reward rather than long-term mission success~\cite{cormen2009introduction}.

\subsection{Monte Carlo Tree Search (MCTS)}

MCTS~\cite{browne2012survey} constructs a search tree from the current mission state, recursively expanding child nodes corresponding to feasible debris selections or refueling decisions. At each node, the Upper Confidence Bound (UCB) formula guides the balance between exploration and exploitation:
\begin{equation}
\text{UCB}_i = Q_i + C \sqrt{\frac{\ln N}{n_i}}
\end{equation}
where $Q_i$ is the mean simulated return for child $i$, $n_i$ is the number of times child $i$ has been visited, $N$ is the total number of simulations from the parent node, and $C$ is the exploration constant. Leaf nodes are expanded using random rollouts to estimate the cumulative reward (e.g., total debris visited), and backpropagation updates values up the tree. The action selected is the child of the root with the highest visit count or value after a fixed number of simulations.

\subsection{Reinforcement Learning Agent}
Proximal Policy Optimization (PPO)~\cite{schulman2017ppo} is a policy gradient algorithm that stabilizes training by constraining policy updates through a clipping mechanism, and in this work we employ a specialized variant, Masked PPO, which extends PPO by incorporating action masks to exclude infeasible actions from the policy distribution~\cite{tang2020ppo,huang2022mask}.
This was used to train a RL agent within a custom OpenAI Gym environment that simulates the ADR mission~\cite{Bandyopadhyay2024revisiting,Bandyopadhyay2025optimizing}. The observation vector $\mathbf{o}_t$ at timestep $t$ includes:
\begin{itemize}
    \item A binary mask of visited debris
    \item Remaining $\Delta V$ budget
    \item Remaining mission time
    \item Current chaser orbital parameters (all six Keplerian elements)
    \item Orbital parameters of all debris (all six Keplerian elements)
\end{itemize}
The discrete action space $\mathcal{A}_t$ consists of all debris plus the option to refuel at the station.
We used action masking to mask the visited debris from the action space over time so that only new debris can be visited over time.The reward function is defined as:
\begin{equation}
r_t = 
\begin{cases}
+1 & \text{for each successful debris rendezvous}\\
0 & \text{for refueling or detour actions}\\
-1 & \text{for early termination or constraint violation}
\end{cases}
\end{equation}
If the agent is runs out of fuel or time the episode is terminated and this makes the agent learn over time to prioritize visiting new debris.
Training is performed with randomly generated debris fields to encourage generalization and robustness across mission scenarios.

\subsection{Unified Evaluation Framework}

All planning algorithms are evaluated within a unified simulation framework, ensuring consistent orbital dynamics, mission constraints, and environment interactions. Key performance metrics include:
\begin{itemize}
    \item \textbf{Debris Visited:} Total number of distinct debris serviced per episode.
    \item \textbf{Total Computational time:} Total time taken by the algorithm to plan the mission.
\end{itemize}
This evaluation protocol allows for direct, fair comparison of solution quality, mission efficiency, and computational tractability across the three planning paradigms.

\section{Results and Discussion}

\begin{figure}[!t]
    \centering
    \includegraphics[width=\linewidth]{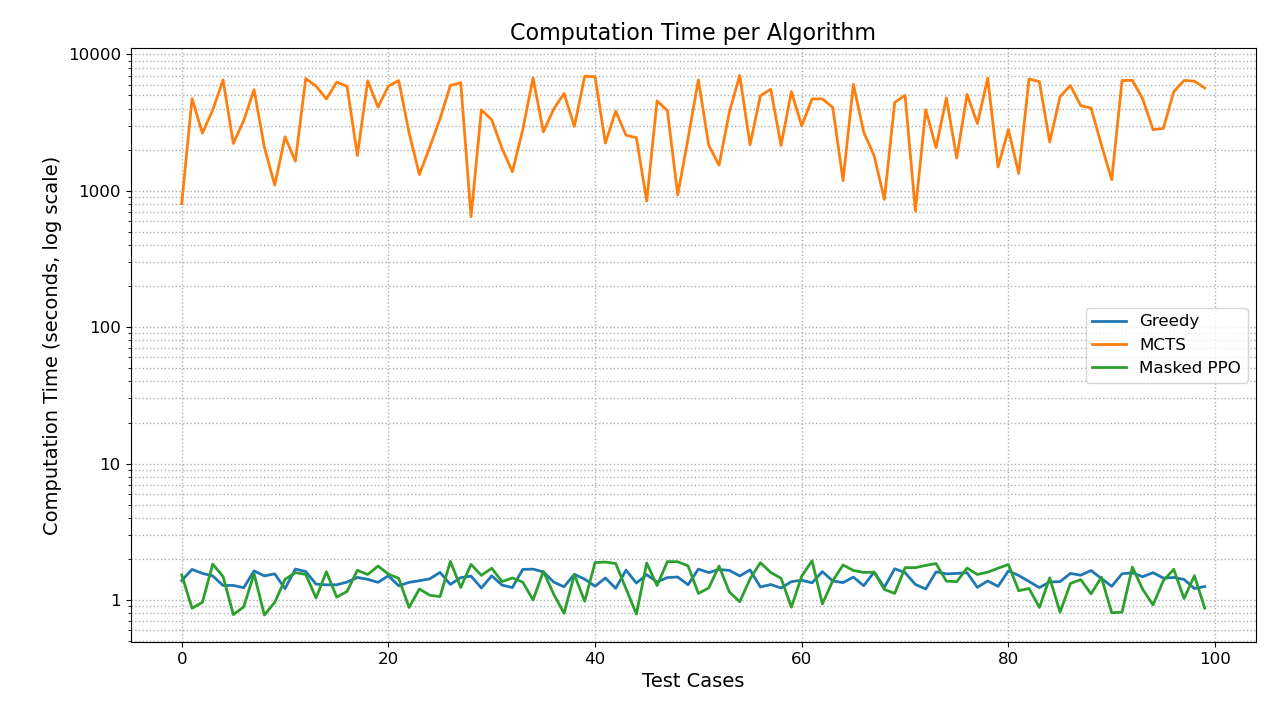}
    \caption{Computation time (log scale) for each algorithm per test case. Masked PPO and Greedy achieve fast runtimes, while MCTS is orders of magnitude slower.}
    \label{fig:computation_algo}
\end{figure}

The performance of the three planning algorithms—Greedy heuristic, Monte Carlo Tree Search (MCTS), and Masked Proximal Policy Optimization (PPO)—was evaluated over 100 randomized test cases. The primary metrics of interest are (i) the total number of debris objects successfully visited per episode and (ii) the computation time required for each planner, as shown in Figures~\ref{fig:debris_per_algo} and~\ref{fig:computation_algo}.

\subsection{Debris Removal Efficiency}

\begin{figure}[!t]
    \centering
    \includegraphics[width=\linewidth]{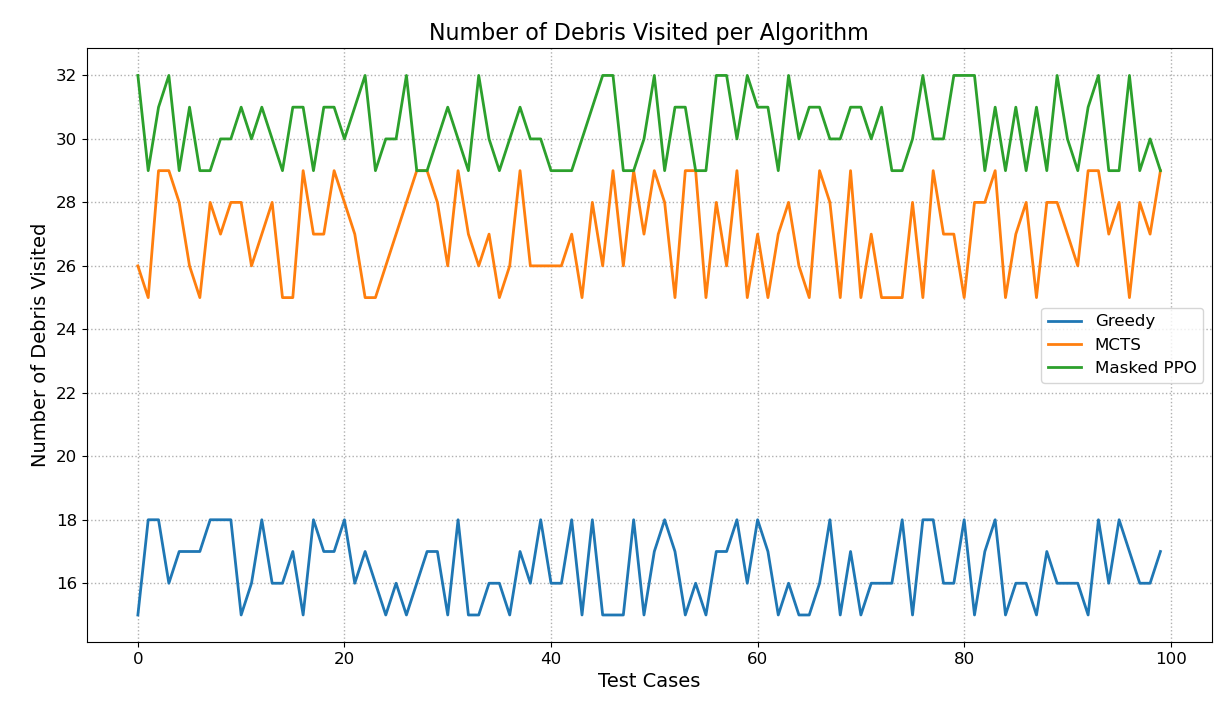}
    \caption{Number of debris objects visited per episode by each algorithm across 100 randomized test cases.}
    \label{fig:debris_per_algo}
\end{figure}

Figure~\ref{fig:debris_per_algo} shows the number of debris visited per test case for each algorithm. The Greedy heuristic consistently achieves the lowest performance, visiting between 15 and 18 debris objects per episode, due to its myopic decision-making and lack of long-term trajectory optimization. MCTS significantly improves upon this baseline, with visit counts ranging from 25 to 29 debris per episode. The improved performance of MCTS is attributed to its ability to simulate future trajectories and optimize cumulative reward, at the expense of higher computational complexity.

The Masked PPO agent outperforms both baselines, visiting between 29 and 32 debris objects per test case. Its ability to generalize across randomized debris fields and to learn long-term reward maximization enables it to consistently select more efficient multi-target rendezvous sequences. The RL agent's performance demonstrates the value of learning-based strategies in high-dimensional, sequential decision-making environments, where both operational constraints and future rewards must be balanced.

\subsection{Computation Time}

Computation time for each algorithm is shown in Figure~\ref{fig:computation_algo} on a logarithmic scale. The Greedy and Masked PPO planners both exhibit extremely fast inference, typically solving each episode in 1–2 seconds on average. In contrast, MCTS is orders of magnitude slower, with episode computation times ranging from approximately 1,000 to 10,000 seconds, due to the need for extensive tree expansion and rollouts at each decision point. This substantial difference highlights a key practical limitation of search-based methods for large-scale, real-time mission planning.

\subsection{Discussion}

The results clearly indicate that Masked PPO achieves the best balance between solution quality and computational efficiency. While MCTS is capable of generating high-quality plans, its prohibitive runtime makes it impractical for on-board or real-time applications. The Greedy agent, though computationally efficient, is severely limited by its short-sightedness, missing many reachable debris objects in complex scenarios.

The Masked PPO agent demonstrates strong generalization to randomized debris fields and robustly respects mission constraints, including refueling logic. These findings support the growing consensus that deep reinforcement learning, when combined with accurate orbital dynamics simulation and carefully designed action masking, can enable scalable and reliable mission planning for active debris removal and similar sequential decision-making problems in space operations.

\section{Conclusion}

This paper presented a unified co-elliptic maneuver framework for multi-target active debris removal (ADR), integrating Hohmann transfers, safety ellipse maneuvers, and explicit refueling logic within a realistic orbital environment. We systematically tested our policy against traditional planning heuristics like—Greedy heuristic and Monte Carlo Tree Search (MCTS)—under identical mission constraints and randomized debris scenarios.

Our results demonstrate that the Masked PPO agent achieves the best overall trade-off between mission efficiency and computational feasibility, consistently outperforming traditional search-based and heuristic methods in both solution quality and runtime. While MCTS approaches near-optimal visitation counts, its high computational cost limits its practicality for real-time or on-board implementation. The Greedy agent offers fast but suboptimal solutions, frequently missing reachable debris due to its myopic strategy.

These findings highlight the potential of modern RL approaches for complex, high-dimensional space mission planning tasks, where long-term objectives, operational safety, and computational efficiency must all be balanced. Future work will explore the integration of more detailed dynamical models, transfer learning for adaptation to changing debris fields, and further onboard validation to support next-generation ADR missions to make the simulation more realistic and practical.

\section*{APPENDIX}
\addcontentsline{toc}{section}{APPENDIX}  

\subsection*{A. Mission Parameters}
\begin{itemize}
    \item \textbf{Initial Orbit and Refuelling Orbit:} 700 km circular orbit, 96° inclination.
    \item \textbf{Target Debris:} 50 objects {uniformed} sampled per episode randomly between 700 and 800 km altitude.
    \item \textbf{Fuel Capacity:} Normalized to 1.0 at start; reset to full on refuel calculated with reference to the maximum dV.
    \item \textbf{Max dV:} Set at 3 km/s.
    \item \textbf{Maximum Episode Duration:} Each test case has a duration of 7 days within which the agent has to visit as many debris as possible.
    \item \textbf{Episode Termination:} All debris visited, fuel exhausted, running out of time or invalid action.
    \item \textbf{Test Case Variation:} Each test case has generates 50 random debris. Each test case is tested for 10 iterations with each strategy as the collisions are completely randomized per iteration.
    \item \textbf{Assumptions:} We assume no effect of the J2 pertubations.

\end{itemize}

\subsection*{B. Reinforcement Learning Hyperparameters}
The hyperparameters were tuned as per our previous work.
\begin{itemize}
    \item \textbf{Algorithm:} Proximal Policy Optimization (Masked PPO).
    \item \textbf{Learning Rate:} $5 \times 10^{-6}$.
    \item \textbf{Training Steps:} 10 million.
    \item \textbf{Batch Size:} 2048.
    \item \textbf{Discount Factor ($\gamma$):} 0.99.
    \item \textbf{Clipping Parameter ($\epsilon$):} 0.2.
    \item \textbf{Action Masking:} Enabled (invalid actions set to $-\infty$ logits).
    \item \textbf{Policy Network:} Multilayer Perceptron (MLP) with two hidden layers, 256 neurons each, \texttt{tanh} activation.
    \item \textbf{Framework:} Stable-Baselines3 + Poliastro.
    
\end{itemize}

\subsection*{C. Plain MCTS and Greedy Configuration}
\begin{itemize}
    \item \textbf{MCTS Algorithm:} Plain Monte Carlo Tree Search using UCT.
    \item \textbf{Exploration Constant:} $c_{\text{uct}} = 1.5$.
    \item \textbf{Simulations per Step:} 200.
    \item \textbf{Rollout Depth:} 15.
    \item \textbf{Greedy Algorithm:} The standard greedy algorithm is used and it chooses the rendezvous with the lowest dv.
    
\end{itemize}

\subsection*{E. Environment Settings}
\begin{itemize}
    \item \textbf{Astrodynamics Libraries:} Poliastro, Astropy.
    \item \textbf{Hardware:} MacBook Pro M1 Max, 64 GB RAM, Python 3.11.
\end{itemize}

\bibliography{ifacconf}

\end{document}